\documentclass[lettersize,journal]{IEEEtran}
\usepackage{amsmath,amsfonts}
\usepackage{algorithmic}
\usepackage{algorithm}
\usepackage{array}
\usepackage[caption=false,font=normalsize,labelfont=sf,textfont=sf]{subfig}
\usepackage{textcomp}
\usepackage{stfloats}
\usepackage{url}
\usepackage{verbatim}
\usepackage{graphicx}
\usepackage{cite}
\begin{document}

\title{Advanced Artificial Intelligence Strategy for Optimizing Urban Rail Network Design using Nature-Inspired Algorithms}

\author{Hariram Sampath Kumar, \IEEEmembership{Member, IEEE,} Archana Singh, Manish Kumar Ojha

\thanks{Hariram Sampath Kumar is with the Department of Artificial Intelligence, Amity University, Noida, Uttar Pradesh - 201313, India (e-mail: harisb4160176@gmail.com).}
\thanks{Archana Singh is with the Department of Artificial Intelligence, Amity University, Noida, Uttar Pradesh - 201313, India (e-mail: asingh27@amity.edu).}
\thanks{Manish Kumar Ojha is with the Department of Artificial Intelligence, Amity University, Noida, Uttar Pradesh - 201313, India (e-mail: mkojha@amity.edu).}}

\maketitle

\begin{abstract}
This study introduces an innovative methodology for the planning of metro network routes within the urban environment of Chennai, Tamil Nadu, India. A comparative analysis of the modified Ant Colony Optimization (ACO) method (previously developed) with recent breakthroughs in nature-inspired algorithms demonstrates the modified ACO’s superiority over modern techniques. By utilizing the modified ACO algorithm, the most efficient routes connecting the origin and destination of the metro route are generated. Additionally, the model is applied to the existing metro network to highlight variations between the model's results and the current network. The Google Maps platform, integrated with Python, handles real-time data, including land utilization, Geographical Information Systems (GIS) data, census information, and points of interest. This processing enables the identification of stops within the city and along the chosen routes. The resulting metro network showcases substantial benefits compared to conventional route planning methods, with noteworthy enhancements in workforce productivity, decreased planning time, and cost-efficiency. This study significantly enhances the efficiency of urban transport systems, specifically in rapidly changing metropolitan settings such as Chennai.

\end{abstract}


\section{Introduction}
Based on the literature survey, a research gap was found for application of Artificial Intelligence (AI) in Transportation Route Planning. In this research, a detailed study is done to explore different applications of AI with Geographical Information Systems (GIS) and variety of nature-inspired algorithms. With reference to the previous work of the current research, focused on developing the modified ACO algorithm \cite{b60}. This algorithm is to be compared with other latest nature-inspired algorithms and show its superiority over others. The urban rail network is to be created for Chennai the capital of Tamil Nadu, India. The origin and destination of the rail network is Tambaram and Shollingnallur,  respectively. Additionally, the model generates results for existing urban rail network in Chennai to show the variation between acquired results and the network.

Further in this section, a comparative analysis of various nature-inspired algorithms is presented and the modified ACO algorithm is chosen by stating the reason for superiority over others. In the upcoming section, a detailed review of various research \& survey papers has been included. In section 3, a detailed methodology of the research and different data utilized for the research has been discussed. In section 4, the results acquired using modified ACO algorithm and analysing different data using Python is presented and discussed. Followed by in the final section, a short brief about the objectives achieved and the future works has been discussed.

\subsection{Comparative Analysis of Nature Inspired Algorithms}

\begin{table}[h]
\caption{Comparative Analysis of Latest Nature-Inspired Algorithms
\label{tab:table1}}
\centering
\resizebox{3.5in}{!}{
\begin{tabular}{|c|p{3cm}|p{2cm}|p{4.5cm}|p{4.5cm}|}
\cline{1-5} 
\hline
\multicolumn{1}{|c|}{\textbf{S. No.}} & \multicolumn{1}{|c|} {\textbf{Article Title and Authors}} & 
\multicolumn{1}{|c|}{\textbf{Algorithm /  Model}} & \multicolumn{1}{|c|}{\textbf{Observations}} & 
\multicolumn{1}{|c|}{\textbf{Challenges}} \\
\hline
\centering 1. & El-Sayed M. El-kenawy, et al., “Greylag Goose Optimization: Nature-inspired optimization algorithm” \cite{b57} & Greylag Goose Optimization Algorithm & 
\begin{itemize}
    \item Balanced exploration \& exploitation.
    \item Easy to implement due to basic concept.
    \item Good convergence speed for certain optimization problems.
\end{itemize}
& 
\begin{itemize}
    \item Tuning of parameters like population size, mutation rate, and crossover rate is a difficult task.
    \item Lack of comprehensive theoretical analysis.
    \item Advanced benchmarking to be used. 
\end{itemize}
\\
\hline
\centering 2. & Marzia Ahmed, et al., “Gooseneck barnacle optimization algorithm: A novel nature inspired optimization theory and application” \cite{b58} & Gooseneck Barnacle Optimization Algorithm & 
\begin{itemize}
    \item Able to maintain population diversity.
    \item Handle large no. of decision variables without computational complexity.
    \item Adaptable to different domain.
\end{itemize}
& 
\begin{itemize}
    \item Converges prematurely to suboptimal solutions.
    \item Only initial success in certain applications.
    \item Will not explore high-dimensional solution spaces.
\end{itemize}
\\
\hline
\centering 3. & Muxuan Han, et al., “Walrus optimizer: A novel nature-inspired metaheuristic algorithm” \cite{b59} & Walrus Optimizer & 
\begin{itemize}
    \item Maintain population of candidate solutions.
    \item Designed for multiple domain problems.
\end{itemize}
& 
\begin{itemize}
    \item Under development, for novelty \& validation.
    \item Lack of comparison with existing algorithms.
\end{itemize}
\\
\hline
\centering 4. & Mohamed Abdel-Basset, et al., “Nutcracker optimizer: A novel nature-inspired metaheuristic algorithm for global optimization and engineering design problems” \cite{b42} & Nutcracker Optimizer &
\begin{itemize}
    \item Adaptable to different field of applications.
    \item Inspired by nutcracker birds for search process.
\end{itemize}
& 
\begin{itemize}
    \item Converges to sub-optimal solution.
    \item Lack of research.
    \item Tough to handle large-scale optimization problems.
\end{itemize}
\\
\hline
\centering 5. & Benyamin Abdollahzadeh, et al., “Mountain Gazelle Optimizer: A new Nature-inspired Metaheuristic Algorithm for Global Optimization Problems” \cite{b41} & Mountain Gazelle Optimizer & 
\begin{itemize}
    \item Explored various optimization problems.
    \item Easy to implement.
\end{itemize}
& 
\begin{itemize}
    \item Lack of theoretical understanding.
    \item Tuning of parameters.
\end{itemize}
\\
\hline
\centering 6. & Farid MiarNaeimi, et al., “Horse herd optimization algorithm: A nature-inspired algorithm for high-dimensional optimization problems” \cite{b27} & Horse Herd Optimization Algorithm & 
\begin{itemize}
    \item Inspired by collective behaviour of herd of horses.
    \item Exhibits collective intelligence during search process.
    \item Dynamically adaptable.
\end{itemize}
& 
\begin{itemize}
    \item Rigorous validation is needed.
    \item Lack of application in broader community.
\end{itemize}
 \\
\hline
\centering 7. & Changting Zhong, et al., “Beluga whale optimization: A novel nature-inspired metaheuristic algorithm” \cite{b36} & Beluga Whale Optimization Algorithm & 
\begin{itemize}
    \item Inspired by social behaviour of beluga whales.
    \item Simulates social interaction for effective collaboration.
    \item Maintained diversity within groups.
\end{itemize}
& 
\begin{itemize}
    \item Lack of theory.
    \item Efficiency not maintained has problem size increased.
    \item Imbalance between exploration \& exploitation.
\end{itemize}
\\
\hline
\centering 8. & Laith Abualigah, et al., “Reptile Search Algorithm (RSA): A nature-inspired meta-heuristic optimizer” \cite{b31} & Reptile Search Optimization Algorithm & 
\begin{itemize}
    \item Simulates foraging behaviour of reptiles.
    \item Adaptable to multiple environments.
\end{itemize}
& 
\begin{itemize}
    \item Lack of theory and practical application.
    \item Tuning of parameters – population size \& exploration rate for optimality.
\end{itemize}
\\
\hline
\centering 9. & N. Eslami, et al., “Aphid–Ant Mutualism: A novel nature-inspired metaheuristic algorithm for solving optimization problems” \cite{b40} & Aphid-Ant Mutualism based Optimization Algorithm & 
\begin{itemize}
    \item Simulates cooperative behaviour between aphids \& ants.
    \item Adaptable and shares information with individuals.
\end{itemize}
& 
\begin{itemize}
    \item Premature convergence.
    \item Lack of validation and needs higher benchmarking.
\end{itemize}
\\
\hline
\centering 10. & Benyamin Abdollahzadeh, et al., “African vultures optimization algorithm: A new nature-inspired metaheuristic algorithm for global optimization problems” \cite{b28} & African Vulture Optimization Algorithm & 
\begin{itemize}
    \item Simulates foraging behaviour of African Vultures.
    \item Combination of local exploration \& long-range movement.
    \item Maintain candidate solution and employability of mechanism.
\end{itemize}
& 
\begin{itemize}
    \item Improper maintenance of efficiency for higher problem size.
    \item Lack of convergence properties and optimization behaviour.
\end{itemize}
\\
\hline
\end{tabular}
}
\end{table}

In Table I, ten different nature-inspired algorithms are compared with respect to their method, applications, benefits, and drawbacks. The modified ACO algorithm (previous works) \cite{b60} is superior than the above-mentioned algorithms due to the following reasons:
\begin{itemize}
    \item Based on the experimental work in our previous work, the modified ACO algorithm showed a better time complexity for the current application compared to other equivalent algorithms.
    \item The scope of research done for above algorithms are to be explored more in different applications and conditions.
    \item The modified ACO algorithm is the best algorithm for optimal route determination problem.
\end{itemize}

\section{Literature Review}
The Preferred Reporting Items for Systematic Reviews and Meta-Analyses (PRISMA) methodology is done to select the important research and review articles based on the scope of the work using pre-defined criteria. It is for faster and in-depth analysis of articles as per the need. A total of 67 articles is reviewed as part of the research and considered 32 articles specifically for the review using the PRISMA. Fig. 1 is a flowchart to depict the entire process of the review.

\begin{figure}[h]
\centering{\includegraphics[width=3.4in]{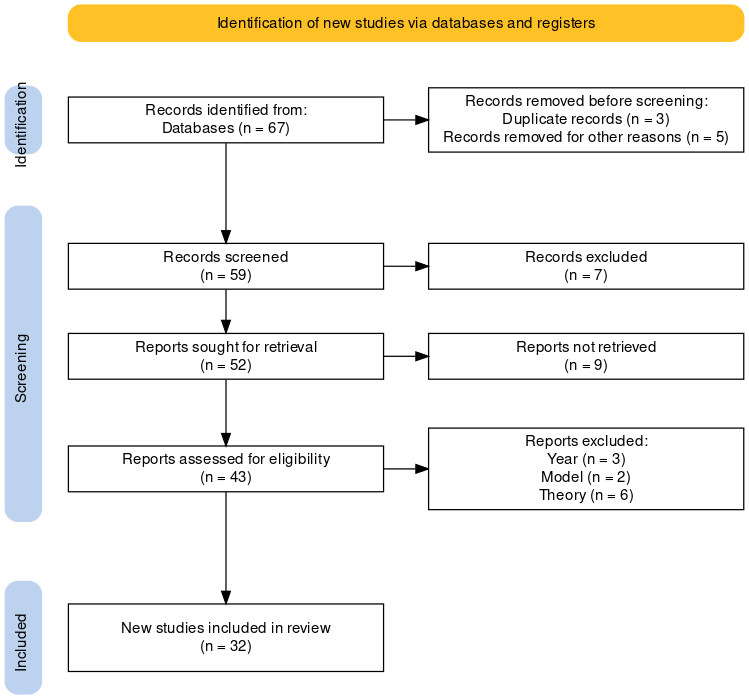}}
\captionsetup{justification=centering}
\caption{PRISMA Flowchart \cite{b61}}
\end{figure}

Ecological and evolutionary studies employ SDMs extensively. Most SDM methods involve considerable GIS data preparation before model creation. This stage is often difficult and might lead to poorly parameterized models or SDM implementation issues. SDM development is not the last outcome for many investigations, and processing after modelling might be difficult. SDMtoolbox streamlines complex species distribution modelling and geospatial analytic pre- and post-processing steps. A single interface houses 59 Python-based GIS utilities in the SDMtoolbox. Several tools improved Maxent SDM prediction performance. SDMtoolbox evaluates haplotype network landscape connectivity using paths or least-cost corridors. It also corrects SDM data overpredictions, quantifies distribution changes between present and future SDMs, and computes biodiversity indicators. SDMtoolbox, for ArcGIS 10.1 or above, focuses on macroecology, landscape genetics, and evolution. Spatial Analyst extension-compatible. The toolbox eliminates post-SDM analysis, redundant \& time-consuming climate data pre-processing, and simplifying species distribution modelling and other GIS analysis chores \cite{b5}.

Particle Swarm Optimisation (PSO) is used to suggest a new urban park spatial location allocation (SLA) method. PSO uses swarm intelligence for efficient optimisation. Population-based random search algorithms based on bird flock social behaviour is used in the study. PSO is simpler, easier to build, and requires less parameters than other AI methods. It considers population density, accessibility, and competitiveness while configuring urban parks. SLA computation using the standard overlaying method is challenging to determine park areas that match these characteristics. Reduce computational complexity and effectively identify parks with Particle Swarm Optimisation (PSO). The service area in urban park simulation findings proved public green-space service fairness. For Service Level Agreement (SLA) issues, Particle Swarm Optimisation (PSO) was found to be practical and effective. The technique can be used in water-saving irrigation systems, farm service centres, hospitals, supermarkets, and cinemas \cite{b6}.

MSW is an increasing problem in towns globally since it contributes to global warming and climate change. Municipal solid waste (MSW) collection and disposal, in developing nations, are essential for environmental and landscape sustainability. These factors greatly affect life quality and expectancy. The inquiry focuses on MSW collection in Danang, one of Vietnam's four major cities. Danang city has a high garbage burden per capita and is affected by climate change, especially extreme weather and natural disasters. A new truck routing model for MSW collection in Danang is proposed. This study uses Chaotic PSO and ArcGIS to optimise Danang's vehicle routing model. Empirical results on Danang's dataset show that the hybrid strategy generates higher total waste quantity than existing methods, including human MSW collection \cite{b7}. The test system simulates dynamic indoor pedestrian evacuation during emergencies. An agent model, floor field model, and PSO are used. The agent-based model simulates individual movement using particle swarm optimisation, whereas the floor field model simulates crowd movement. The software has three primary modules: 3D visualisation, fire spread simulation, and pedestrian evacuation movement simulation. The software provides a complete feature for public security research and emergency management \cite{b8}.

The visually handicapped have a significant obstacle when it comes to exploring new surroundings. Therefore, there is a want for a system that can provide safe assistance to individuals during their travel. The present research provides a novel approach to path planning by utilising the Particle Swarm Optimisation (PSO) algorithm, incorporating specified waypoints. The aforementioned method calculates the shortest possible route from the starting place to the ending position, based on the predetermined waypoints. The provision of an accessible pedestrian route for those with visual impairments can be achieved by the utilisation of predetermined waypoints (set of coordinates along pedestrian walkaway). The system comprises a process for selecting a destination and a process for designing a journey. During the destination selection process, users are provided with a comprehensive list of possible sites from which they can choose the specific locations they wish to visit. The path planning procedure utilises the Euclidean distance formula to establish the path length between all nodes, taking into account preset waypoints, the start points, and the final destination. The Particle Swarm Optimisation (PSO) algorithm aims to minimise the overall cost for path length in order to find the shortest possible route. The simulation analysis conducted on the proposed strategy has yielded encouraging outcomes in terms of determining the ideal path for various destinations. The findings of this study will be utilised to investigate the feasibility of creating a path guidance system for those with visual impairments, enabling them to navigate unfamiliar surroundings autonomously \cite{b10}.

EvoloPy is a Python framework that is both open source and cross-platform. It is designed to implement a diverse set of metaheuristic algorithms, both classical and contemporary, that draw inspiration from nature. The objective of this model is to streamline the utilisation of metaheuristic algorithms by individuals without specialised knowledge from various disciplines. EvoloPy offers a user-friendly interface and minimal dependencies, facilitating the utilisation of this software by researchers and practitioners. It enables them to optimise and benchmark their specific issues by employing robust metaheuristic optimizers available in the existing literature. The aforementioned framework enables the development of novel algorithms or the enhancement, integration, and evaluation of existing ones \cite{b11}.

Smart cities have the potential to enhance the overall quality of life for its residents, with Intelligent Transport Systems emerging as a prominent subject within this domain. As the population density within a given region continues to rise, urban areas encounter challenges such as persistent traffic congestion. The present research presents an examination of the applications of computational intelligent techniques and analysis in facilitating traffic solutions. In order to enhance the autonomy and intelligence of an environment, the utilisation of prediction models and heuristics is deemed the most effective approach. It presents the introduction of an application that utilises ML and optimisation techniques to improve the capabilities of a smart ecosystem. In order to verify its accuracy, a comprehensive assessment was conducted utilising Multi-Layer Perceptron in conjunction with Particle Swarm Optimisation, wherein it was compared to the most advanced methods now available. The evaluations were conducted utilising authentic data traffic under a situation of unrestricted traffic flow \cite{b15}.

Internet and photo editing apps have undermined multimedia data security in recent years. This work uses Optimal SVD-DWT-based watermarking to prevent unauthorised duplication, and protect copyright. A survey evaluates PSO, Genetic Algorithm, and ABC optimisation techniques to improve watermarking algorithms' robustness and imperceptibility. PSNR, SSIM, and NCC were used in this study as performance measurements. We utilise Python Anaconda for experiments. Additionally, each optimisation algorithm's computing time is assessed and compared. DWT-SVD hashing prevents false positives in SVD-based watermarking through authentication \cite{b22}.

The World Health Organisation advises individuals to engage in a minimum of 8000 steps of daily walking, which can be effectively achieved through hiking. Moreover, it is widely acknowledged that hiking is inherently environmentally sustainable, as engaging in hikes to the closest waterfall or mountain peak not only promotes personal well-being but also contributes to the preservation of the natural resource. By possessing the knowledge of selecting the nearest lodge to the intended destination, they are afforded the most expedient means of embarking on an expedition. These assessments are performed on the waterfalls' heights, the peaks' elevation, and their proximity to the mountain lodges \cite{b24}. Enhancing public transit to minimise journey time and increase access to underserved areas will encourage private vehicle owners to use it. Due to fewer exhaust emissions, reducing vehicle traffic should reduce traffic and air pollution. This study employed GIS, PSO, and GA to better assess Amman bus stop travel time and serviceability. GIS modelling shortened trip time on streets with redundant bus stops at irregular distances. Travel time on Zahran Street dropped 23.25\% off-peak and 19.74\% peak. Compare that to 28.95\% and 39\% reductions from PSO. The introduction of GA reduced Zahran Street travel time by 47.96\%. The addition of bus stops on streets with few stops increased journey time. The increase in stations from 6 to 25 on Al-Quds Street increased travel time. The walking distance to bus stop dropped from over 2000 m to 400 m due to this longer travel duration. Bus stop serviceability is more important than capacity. The mean demand per Al-Quds Street bus stop was 10,456, whereas capacity was 1465. The GA and PSO algorithm are easy to implement in current networks and planned projects, making them user-friendly for urban planning \cite{b25}.

The implementation of emergency evacuation measures during and following a flood is of utmost importance in order to minimise immediate consequences and enhance social resilience for sustained recovery. In order to optimise the evacuation procedure, it is imperative to ascertain and forecast secure regions prior to the occurrence of a flood. The safe area or shelter in place serves a dual purpose during a flood event, functioning as both temporary shelters and meeting sites for individuals to congregate prior to evacuation. The objective of the study is to ascertain the safe zone based on the spatial and environmental attributes of the metropolitan region. The important contribution of this study involves the utilisation of MPSO in conjunction with local search (LMPSO) to identify secure regions. The method under consideration identifies the most advantageous placement of temporary shelters in the form of evacuation stations. The comparative analysis of the outcomes obtained from MPSO and LMPSO revealed that LMPSO exhibits superior efficiency compared to the modified version. LMPSO ensures a balanced distribution of optimal locations for safe areas, hence benefiting the entire population \cite{b30}.

Solar and other renewable energy sources are popular replacements to fossil fuels. The first step to expand solar energy usage is finding good locations for solar power facilities. This study compares PSO and decision tree method for solar power site selection. MCDM is used in decision tree and PSO algorithms to extract high-potential zones for optimal sites. The findings suggest that research region has the capacity to accommodate the establishment of solar power plants, as determined by a three-tier assessment of appropriateness. The optimisation approaches were compared, and it was found that the decision tree had a prediction rate of 0.29 for the extremely high potential class, while the PSO had a prediction rate of 0.13. Hence, the use of the decision tree methodology presents a more effective approach in identifying prospective regions for the advancement of solar energy inside the designated study area. The research findings offer a valuable asset for strategizing and making informed choices on the advancement of sustainable energy in the coming years. The decision tree employed for current study has the potential to be applied in many global regions for the purpose of identifying the most suitable sites for the establishment of renewable energy facilities \cite{b32}.

The computing time and other factors of Python and Golang are examined for naturally inspired computer methods. This research will assess which publicly available platform, Python or Golang, is best for Nature-Inspired algorithm productivity and performance. Nature-inspired algorithms consume a lot of computer power and are easy to implement. They will compare Python and Golang compute times for four Nature-Inspired algorithms \cite{b33}. The performance challenges encountered by particle swarm optimisation (PSO) algorithms mostly revolve around enhancing the quality of solutions, increasing computing speed, optimising computational resources, and addressing large problems. The SMO algorithm operates on the premise of partitioning swarm into subgroups, potentially resulting in a reduction in its speedup. The research introduces a novel approach called multi-swarm-based parallel SMO to address complex problems on a large scale. The implementation of PSMO-MS in this study utilises a synchronous master/slave parallel architecture \cite{b38}.

Due to the difficulty of adequately representing hundreds of lakes, hydrological modelling frequently ignores lake effects or simulates only the most important lakes in a watershed. The paper explores about BasinMaker, a GIS toolset. BasinMaker creates vector-based hydrological routing networks for several rivers and lakes. Hydrological routing models need sub-basin \& lake geometry, network topology, and channel parameters. The HydroLAKES dataset encompasses all lakes with a size above 10 hectares. BasinMaker possesses a distinctive characteristic in its lake representation capabilities, rendering it particularly advantageous for modellers who require explicit representation of several lakes inside their watershed simulation models \cite{b45}.

Walking is a readily accessible and straightforward kind of physical activity that has been shown to be successful in mitigating pollution, conserving resources, and improving the overall well-being of the populace. Prior studies have demonstrated the necessity of developing a wireless interface (WI) that incorporates real-time conditions. The aim of the work was to develop a Geographic Information System algorithm that could automate the computation of real-time and personalised walking intensity (WI) and propose a pleasant walking route based on these inputs. The study took into account two comfort aspects, namely shaded locations and temperature. In order to achieve this objective, we developed an innovative technique for projecting shadows onto the ground and a fresh approach to establishing a network of pavement pathways. An algorithm was developed to effectively calculate WI values while considering the user's preferences. Subsequently, these numbers were utilised to generate the optimal path for the user to traverse the pavements, taking into account the current weather conditions. The key findings can be examined from two perspectives. First, approaches for constructing a network for pedestrians along pavements, traffic crossings and computing building shadows improve GIS functionality. Thus, this would increase walk frequency and duration, improving population health. The validation process involved evaluating the model's accuracy in shadow calculations and analysing the routes generated under various scenarios \cite{b47}.

GIS is the most used technology for mapping \& identifying Groundwater Potential Zones. ML algorithms for replicating GWPZ's influencing parameters may improve model performance. The Karaj-Tehran plain, 5156 square kilometres, in Iran, hosted the research. This plain has flat and hilly topography between 818 and 4374 metres. 75\% of 105 wells were used for model training and 25\% for validation. The study also used SVM, XGB, LR, and RF models to compute influential factor and create GWPZ maps. XGB and RF had greater Kappa coefficients (80\% and 65\%). Performance of tree-based models (XGB and RF) was better than others. The model outputs show that height and slope affected GWPZ projections in the research area. The results showed that ML and GIS accurately identify GWPZ \cite{b49}.

The study optimises the high-pressure H-NG pipeline network with a statistical model. The compressor \& pipeline station model uses gas hydrodynamics. The essay optimises various goals using ant colony optimisation. First, a single-objective optimisation problem determines the best hydrogen-NG mix within pipeline constraints. Operational variables for hydrogen injection into pipeline networks of natural gas at various levels were also investigated. The current study introduces an ACO-optimized multi-objective optimisation model for bi- and tri-objective problems. Bi-objective optimisation has received less attention than single-objective optimisation. Few studies have used tri-objective function assessments to show how hydrogen affects the natural gas (NG) network. This study uses tri-objective \& bi-objective functions to evaluate how hydrogen injection affects operational metrics. The article gives tri-objective \& bi-objective equations for modelling the pipeline network of H-NG. These equations are optimised using ACO to close the gap. Pareto fronts show how various issue objectives trade off. The tri-objective \& bi-objective optimisation tasks aim to maximise natural gas hydrogen mole percent. Optimising delivery station pressure, throughput, and power and minimising compressor fuel use are other goals. The findings will help pipeline operators transport H-NG blend while reducing carbon intensity per unit of energy-delivered fuel in natural gas pipeline networks \cite{b50}.

In recent times, the optimum control problem has garnered significant attention due to its relevance in addressing real difficulties. In this context, meta-heuristic algorithms have demonstrated efficacy in successfully and efficiently tackling these challenges. Nevertheless, it is important that these methods may not be universally applicable in addressing all optimisation problems in accordance with the no free lunch theorem. Hence, there exists a perpetual opportunity for the advancement of novel meta-heuristic algorithms. The current research introduces an optimisation method called as the Tyrannosaurus (T-Rex) Optimisation Algorithm (TROA), which is based on hunting techniques. The algorithm is developed based on the hunting behaviour exhibited by the T-Rex. A total of 12 benchmark problems and 4 practical optimum control issues were utilised to evaluate the performance of this technique. The outcomes achieved by the proposed methodology have demonstrated superior performance in comparison to the aforementioned methodologies \cite{b51}.

GKS optimizer (GKSO) is a revolutionary metaheuristic algorithm (MA) based on Genghis Khan shark behaviour. GKSO is designed for numerical optimisation and engineering design. GKS predation and survival behaviour create GKSO. The optimisation procedure simulates GKS's hunting, movement, foraging, and self-protection. These operators are mocked efficiently with mathematical models to optimize agents in different search space regions. The technique is tested qualitatively and quantitatively on the suggested GKSO to prove its efficacy and superiority. Qualitative research shows GKSO's exploration and exploitation potential. On CEC2019 and CEC2022 datasets, it quantifies GKSO utilising eight fish optimisation methods and nine Mas. The study evaluates GKSO's applicability and resilience through several experiments. These scenarios test GKSO for CEC2022 across maximum fitness evaluation amounts and dimensions. The statistical results show that GKSO outperforms two algorithmic techniques. GKSO and 7 other optimizers are also tested using five constrained optimisation problems (OPs) from the CEC2020 benchmark restricted optimisation functions. The 50 engineering case studies in these OPs demonstrate GKSO's versatility in tackling complicated practical challenges \cite{b52}.

In addition to hydrogen generation, transportation plays a role in the successful establishment of the decarbonised hydrogen economy globally. In order to address the inquiry regarding the production locations, routes, and forms of transportation for the necessary quantities of hydrogen, it is imperative to build a comprehensive model of a sector-integrated hydrogen transport system. The article presents a versatile approach and instrument for the spatial organisation of integrated energy transport systems across different sectors. It allows variable geographical adjustment and conformance to existing energy system models. A unique energy transport system that investigates hydrogen starting point and transit in a hydrogen economy using ship transport and pipeline. The model is validated by comparing it to hydrogen transport flux orientation investigations \cite{b54}.

\section{Methodology}
The objective of the study is proposed in accomplishing the following targets:
\begin{itemize}
    \item Detailed review of latest nature-inspired algorithms and comparative analysis of the same with modified ACO algorithm, to show the superiority of modified ACO algorithm \cite{b60} compared to others.
    \item Validate the model by comparing its result with existing urban rail route in the city.
    \item Optimal route identification for urban rail planning between chosen origin and destination in a city.
    \item Determination of rail stops along the chosen route by considering land usage pattern, geographical data, census data, and point-of-interest places of the city with proposed rules \& conditions.
    \item Creating final route with these stops as intermediate points along the urban rail route.
    \item Give an insight about the future developments of the current research.
\end{itemize}

First, research and review papers are collected from Elsevier, IEEE, Springer, and other databases. A brief study on 67 papers is done and further 32 papers are chosen based on pre-defined criteria like published year, theory, and model used using PRISMA model.

Second, a detailed study was performed on 32 papers and presented a comparative analysis of latest nature-inspired algorithms with modified ACO algorithm \cite{b60} (previous work). Following the analysis, the modified ACO algorithm is superior compared to other algorithms.

Third, the city for urban rail network is chosen.
The developed model is used to find the route for existing urban rail network in the city to show the variations between the original rail route and the route generated by the model. 

Fourth, the origin \& the destination of the new urban rail route is chosen. Further, the land usage data (refer Fig. 2), geographical data (refer Fig. 3), census data (refer Fig. 4), and point-of-interest places (refer Fig. 5) of the city is used to determine the stops for urban rail network.

\begin{figure}[h]
\centering{\includegraphics[width=3.5in]{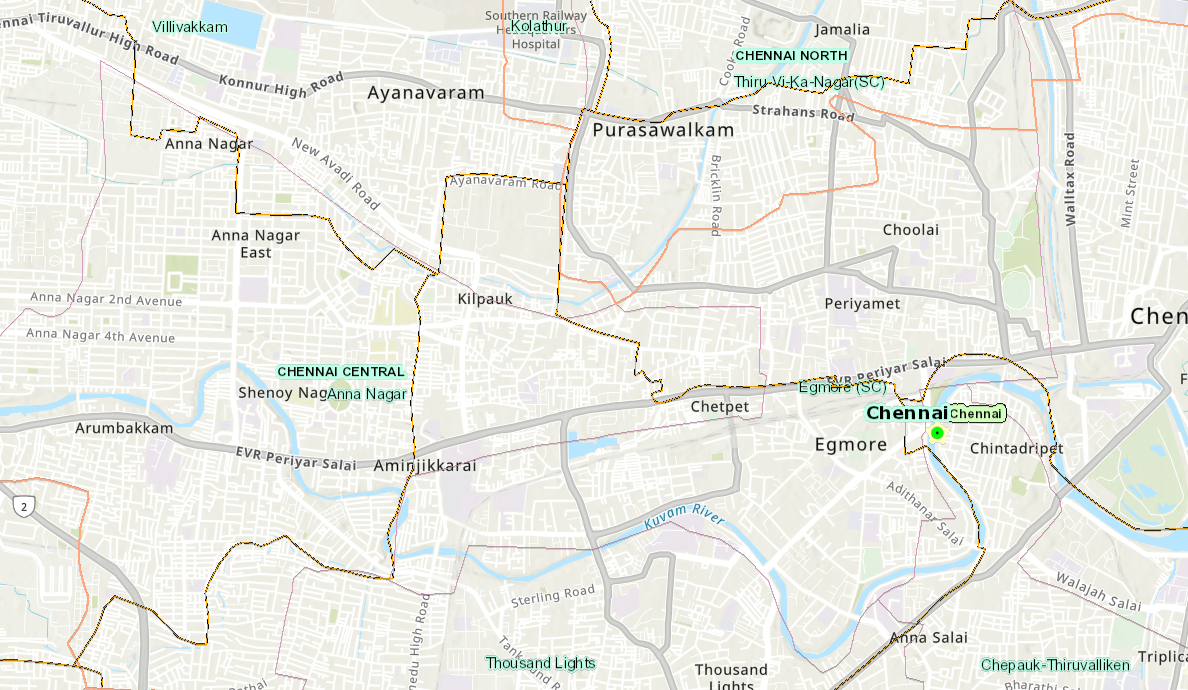}}
\captionsetup{justification=centering}
\caption{Land Usage Data of Chennai City}
\end{figure}

\begin{figure}[h]
\centering{\includegraphics[width=3in, height=2.5in]{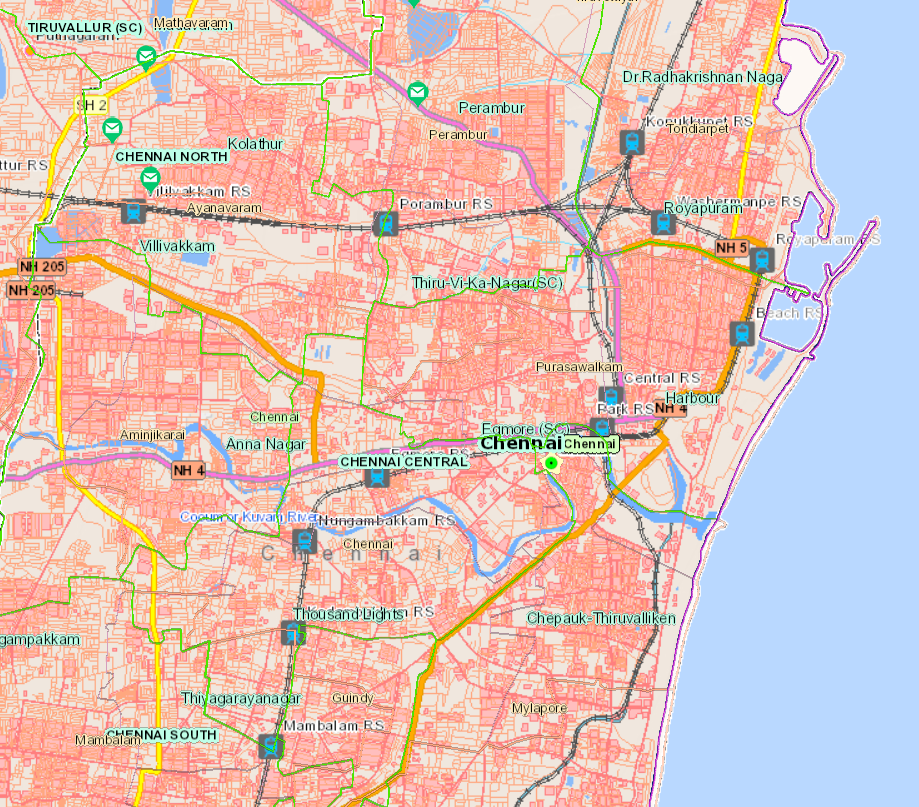}}
\captionsetup{justification=centering}
\caption{Geographical Data of Chennai City}
\end{figure}

\begin{figure}[h]
\centering{\includegraphics[width=3.5in, height=0.5in]{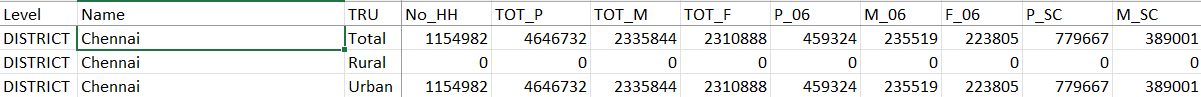}}
\captionsetup{justification=centering}
\caption{Census Data of Chennai City}
\end{figure}

\begin{figure}[h]
\centering{\includegraphics[width=3in, height=2.5in]{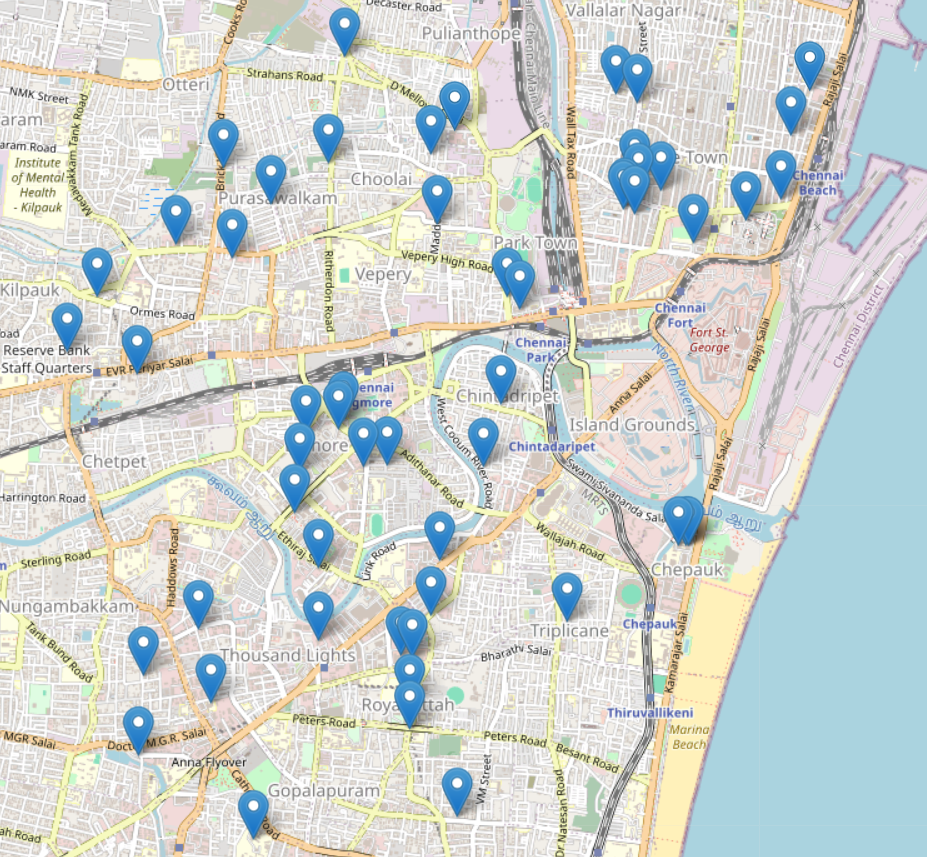}}
\captionsetup{justification=centering}
\caption{Point-of-interest places in Chennai City}
\end{figure}

Fourth, the multiple routes between the origin \& destination are generated and optimal route between origin and destination is chosen by utilising modified ACO algorithm. The final network is created by integrating the determined stops along the chosen route.

The above implementation and determination of urban rail stops is done using Jupyter Notebook (The Python Environment) and Google MAPS Platform. The results are observed for different stages of implementation. The complete planning of the research is represented in the Fig. 6.

\begin{figure}[h]
\centering{\includegraphics[width=3.5in, height=4.75in]{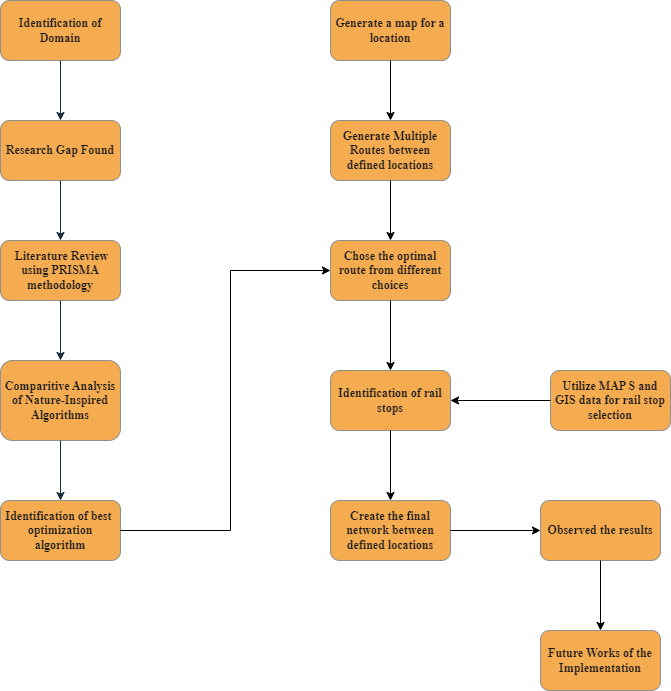}}
\captionsetup{justification=centering}
\caption{Research Framework}
\end{figure}

In Fig. 6, the methodology establishes a framework for optimizing metro network route planning within a specific urban environment. 

The proposed model offers several advantages that can significantly improve urban living. One of the key contribution lies on the potential to enhance efficiency and cost-effectiveness. The model aims to identify the most efficient routes for metro networks, leading to shorter travel times. Additionally, the optimized route planning process could minimize construction expenses compared to conventional methods, leading to cost savings during the development phase. By factoring in real-time data, the model strategically position’s stations to serve the most populated areas and major destinations. This improved accessibility encourages ridership on the metro network. This shift towards public transportation can contribute to a more sustainable transportation system, reducing traffic congestion and its associated environmental impact. With this technology, urban planners can make informed decisions during future infrastructure projects, ensuring optimal resource allocation for metro network development.

By prioritizing efficiency, accessibility, and cost-effectiveness, the proposed model has the potential to contribute to a more sustainable, well-connected, and economically vibrant society.

\section{Result and Discussion}

The recently developed nature-inspired algorithms are compared with modified ACO algorithm of my previous work and the modified ACO algorithm \cite{b60} is considered over other algorithms for choosing optimal route from multiple choices. The algorithm chooses the optimal solution based on total distance and total time taken to travel from given origin to destination. The model developed in the current research is used to create a metro rail route in the city – Chennai, Tamil Nadu. 

The algorithm as determined the route with stops for the existing Chennai Metro Rail Network between Chennai Airport and Thiruvottriyur. The final route is predicted as same as the existing one with a slight variation in number of stops from 14 to 12. The actual metro rail network and step by step implementation of the same using the algorithm is given in Fig. 7, 8, 9, 10, 11.

\begin{figure}[h]
\centering{\includegraphics[width=3in, height=3in]{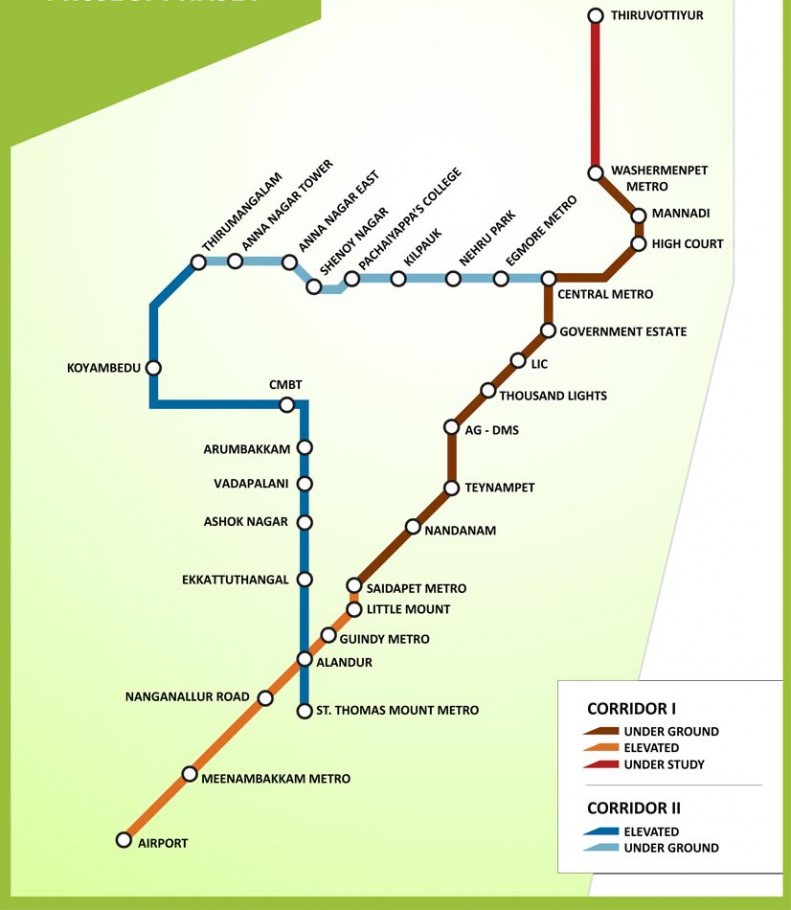}}
\captionsetup{justification=centering}
\caption{Original Route of Chennai Metro Rail Network between Chennai Airport and Thiruvottriyur}
\end{figure}

\begin{figure}[h]
\centering{\includegraphics[width=3in, height=3in]{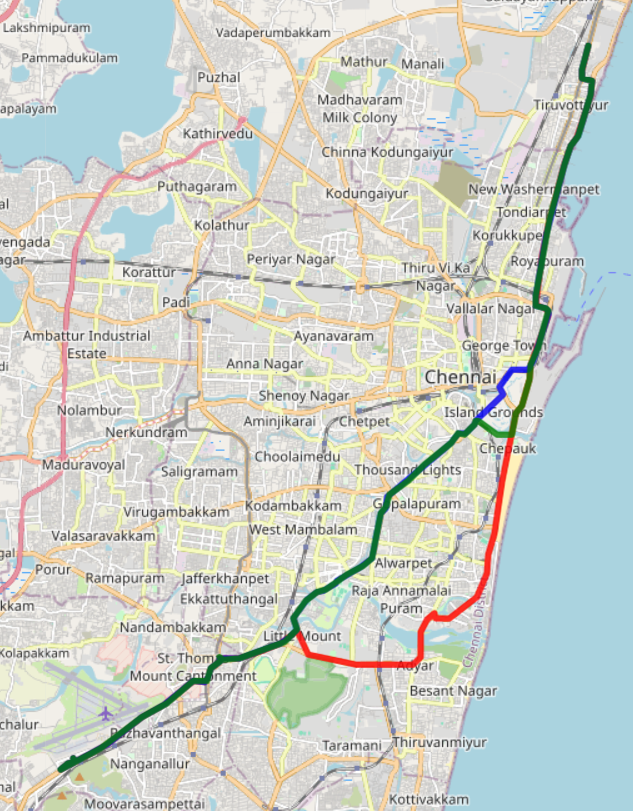}}
\captionsetup{justification=centering}
\caption{Multiple routes generated between Chennai Airport and Thiruvottriyur using developed model}
\end{figure}

\begin{figure}[h]
\centering{\includegraphics[width=3in, height=3in]{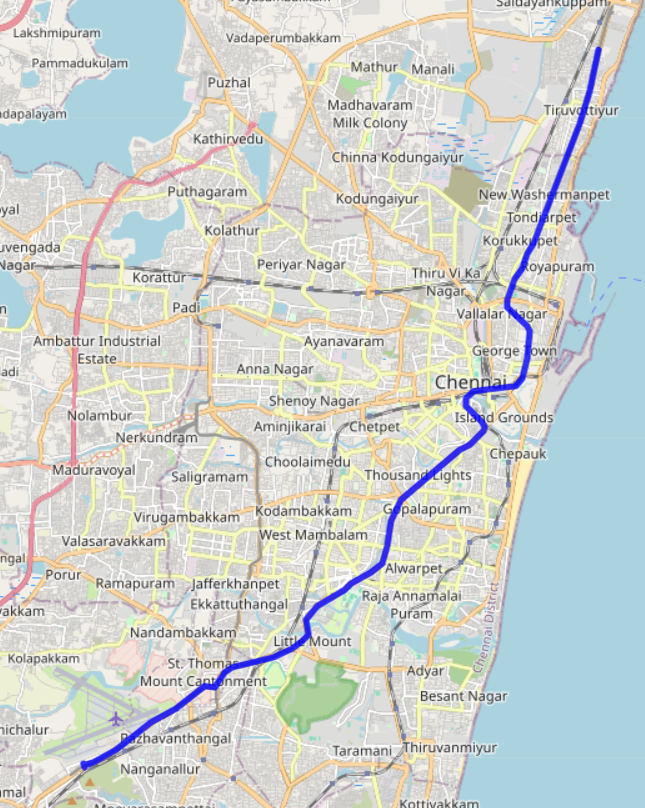}}
\captionsetup{justification=centering}
\caption{Optimal route chosen between Chennai Airport and Thiruvottriyur using developed model}
\end{figure}

\begin{figure}[h]
\centering{\includegraphics[width=3.5in]{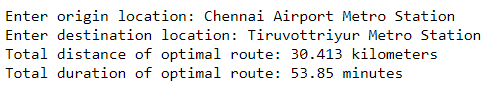}}
\captionsetup{justification=centering}
\caption{Factors influenced the optimal route selection using modified ACO algorithm}
\end{figure}

\begin{figure}[h]
\centering{\includegraphics[width=3in, height=3in]{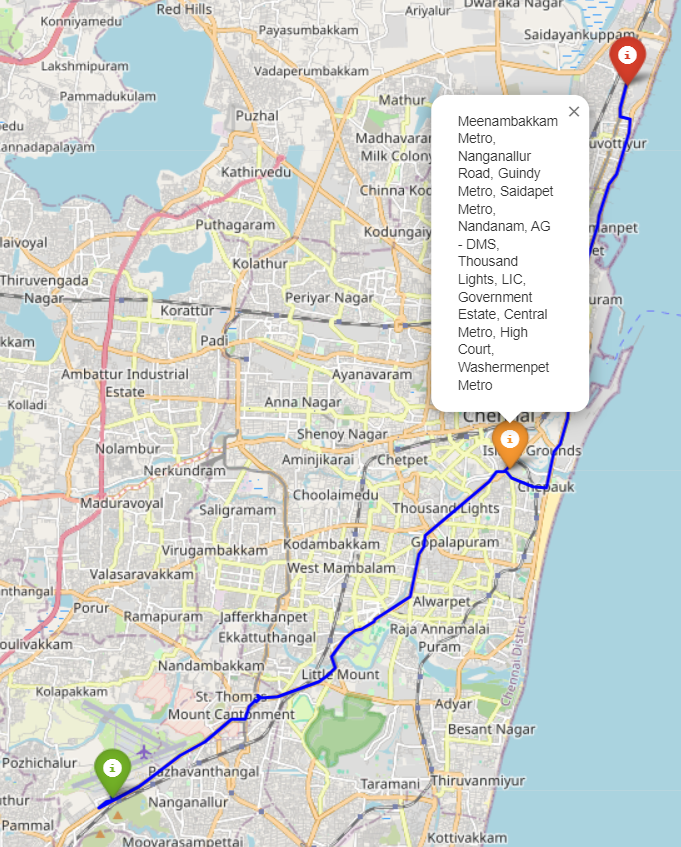}}
\captionsetup{justification=centering}
\caption{Final route between Chennai Airport and Thiruvottriyur with Intermediate stops}
\end{figure}

The origin and destination of the new metro network in the city is Tambaram and Sholingnallur, respectively. The final route between these locations consists of 7 stops at different locations. The actual benefits of the developed model will reduce the work force, time taken, and cost involved for metro route planning in a city (refer Table II).

\begin{table}[h]
\caption{Improvement in Factors Affecting Metro Route Planning
\label{tab:table1}}
\centering
\resizebox{3.5in}{!}{
\begin{tabular}{|c|p{2cm}|p{5cm}|p{5cm}|}
\cline{1-4} 
\hline
\multicolumn{1}{|c|}{\textbf{S. No.}} & \multicolumn{1}{|c|} {\textbf{Factors}} & 
\multicolumn{1}{|c|}{\textbf{Results using Traditional Method}} & \multicolumn{1}{|c|}{\textbf{Results using Developed AI Model}} \\
\hline
\centering 1. & \centering Work Force & 90 to 95 persons &
\textbf{15 to 18 persons}
\\
\hline
\centering 2. & \centering Time Taken & 96 to 108 weeks &
\textbf{5 to 7 weeks}
\\
\hline
\centering 3. & \centering Cost & 250 to 300 crores (INR) & 
\textbf{12 to 15 crores (INR)}
\\
\hline
\end{tabular}
}
\end{table}

The results of different parts of implementation for new urban rail route are presented and explained below:

The Chennai city is found using GIS data and visualized the entire area of the city, as shown in Fig. 12.

\begin{figure}[h]
\centering{\includegraphics[width=3.5in, height=2.5in]{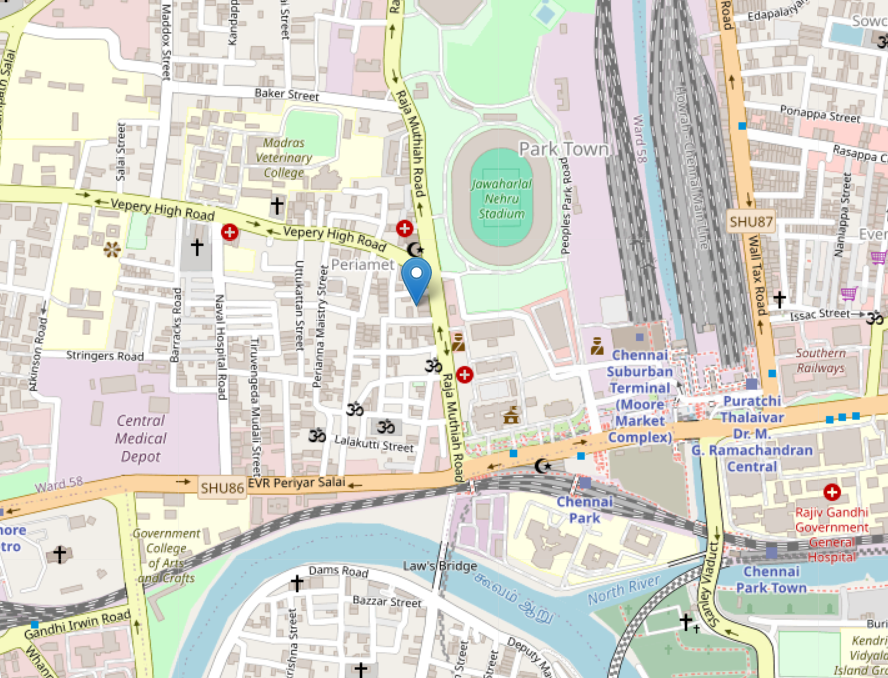}}
\captionsetup{justification=centering}
\caption{Visualization of the Chennai City}
\end{figure}

The Fig. 13, shows the visualization of multiple routes generated between the chosen origin and destination. Followed by the optimal route is selected using the modified ACO algorithm and the selection is influenced by total distance and total time taken. The visualization and results of using the modified ACO algorithm is shown in Fig. 14 and Fig. 15.

\begin{figure}[h]
\centering{\includegraphics[width=3.5in]{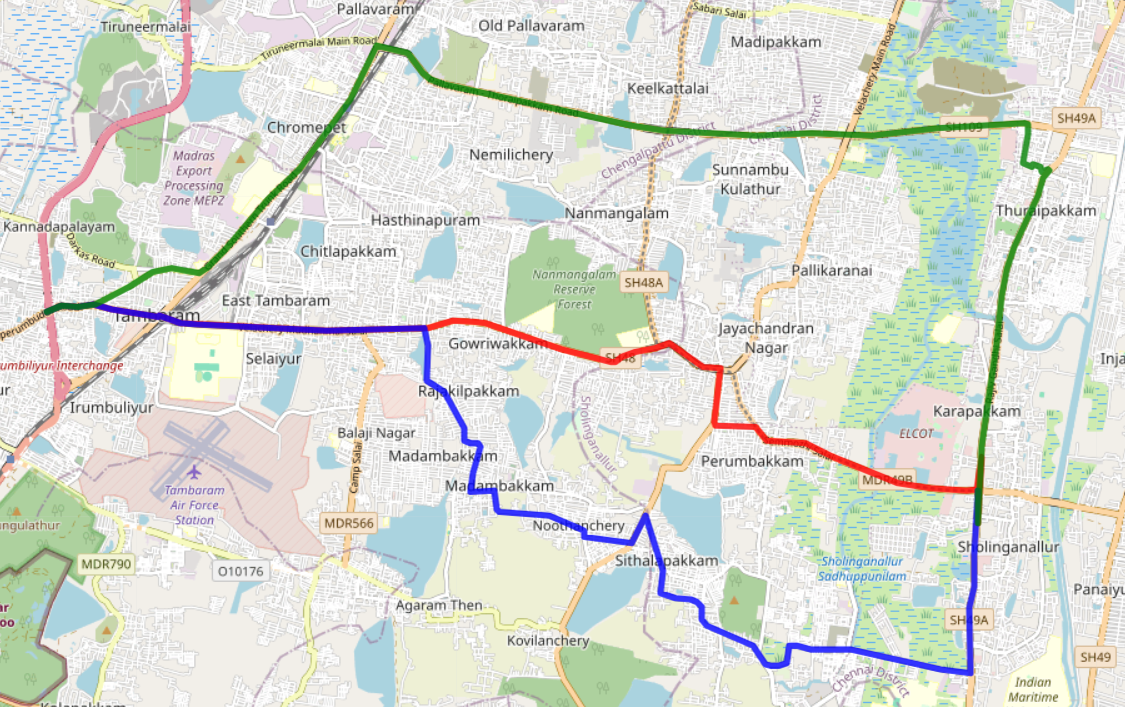}}
\captionsetup{justification=centering}
\caption{Multiple routes generated between Tambaram and Sholingnallur}
\end{figure}

\begin{figure}[h]
\centering{\includegraphics[width=3.5in]{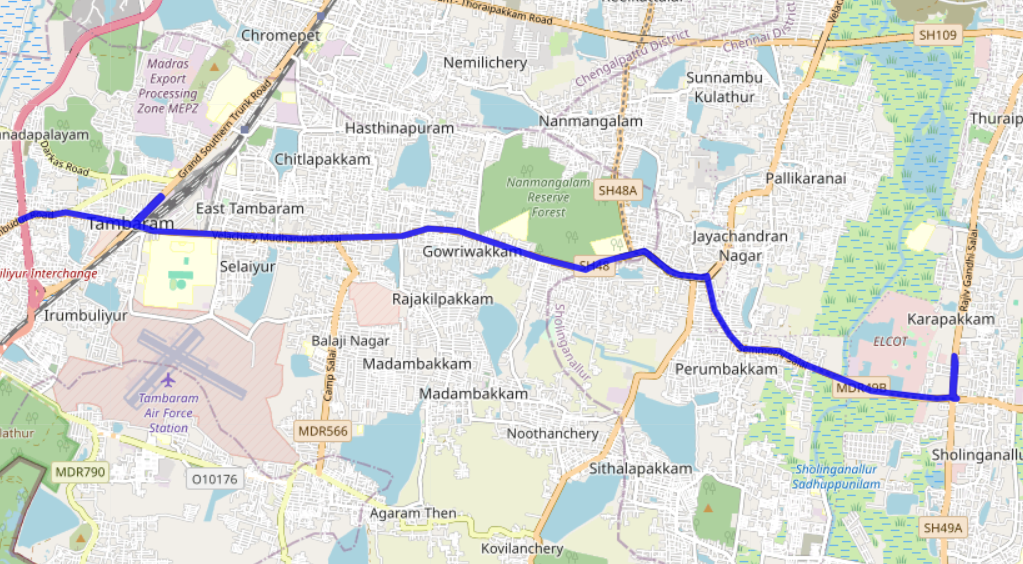}}
\captionsetup{justification=centering}
\caption{Optimal route chosen between Tambaram and Sholingnallur}
\end{figure}

\begin{figure}[h]
\centering{\includegraphics[width=3.5in]{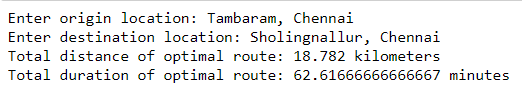}}
\captionsetup{justification=centering}
\caption{Factors influenced the selection of optimal route using modified ACO algorithm}
\end{figure}

The urban rail stops within the Chennai city is determined based on external data like land usage data, census data, point-of-interest places, \& GIS data and by considering few rules \& conditions the locations are determined and visualized, as shown in Fig. 16.

\begin{figure}[h]
\centering{\includegraphics[width=3in, height=2.5in]{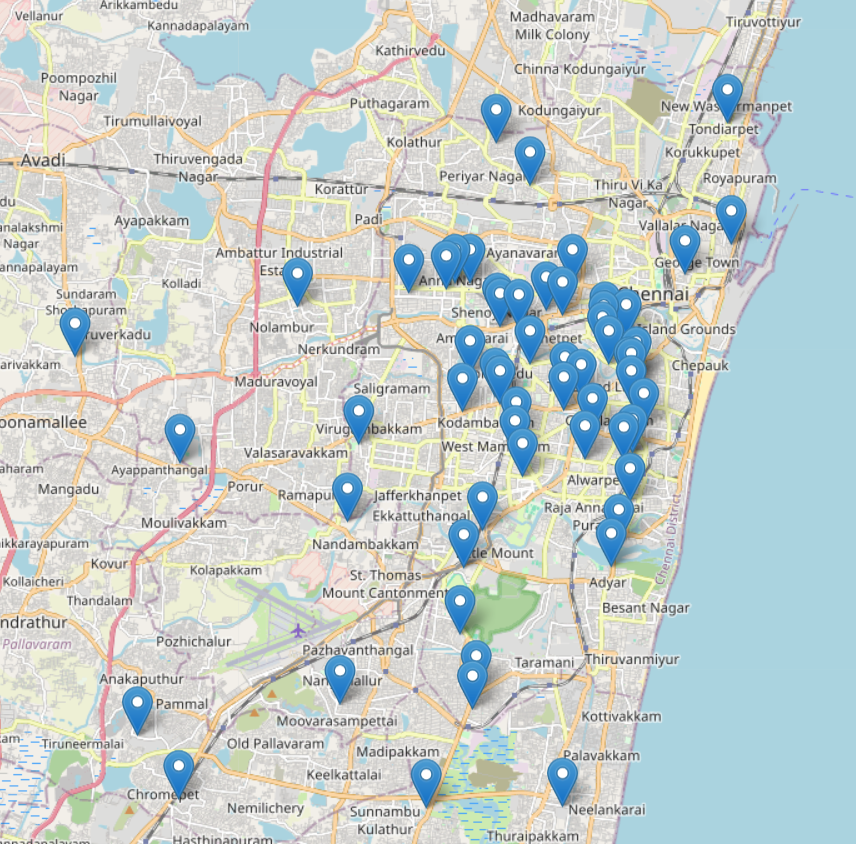}}
\captionsetup{justification=centering}
\caption{All possible stops in the Chennai city for urban rail network}
\end{figure}

The final route between Tambaram and Sholingnallur with intermediate stops along the route is found by considering the nearest determined stops along the route and within a prescribed range of distance. The visualization of the final route is shown in Fig. 17.

\begin{figure}[h]
\centering{\includegraphics[width=3.5in]{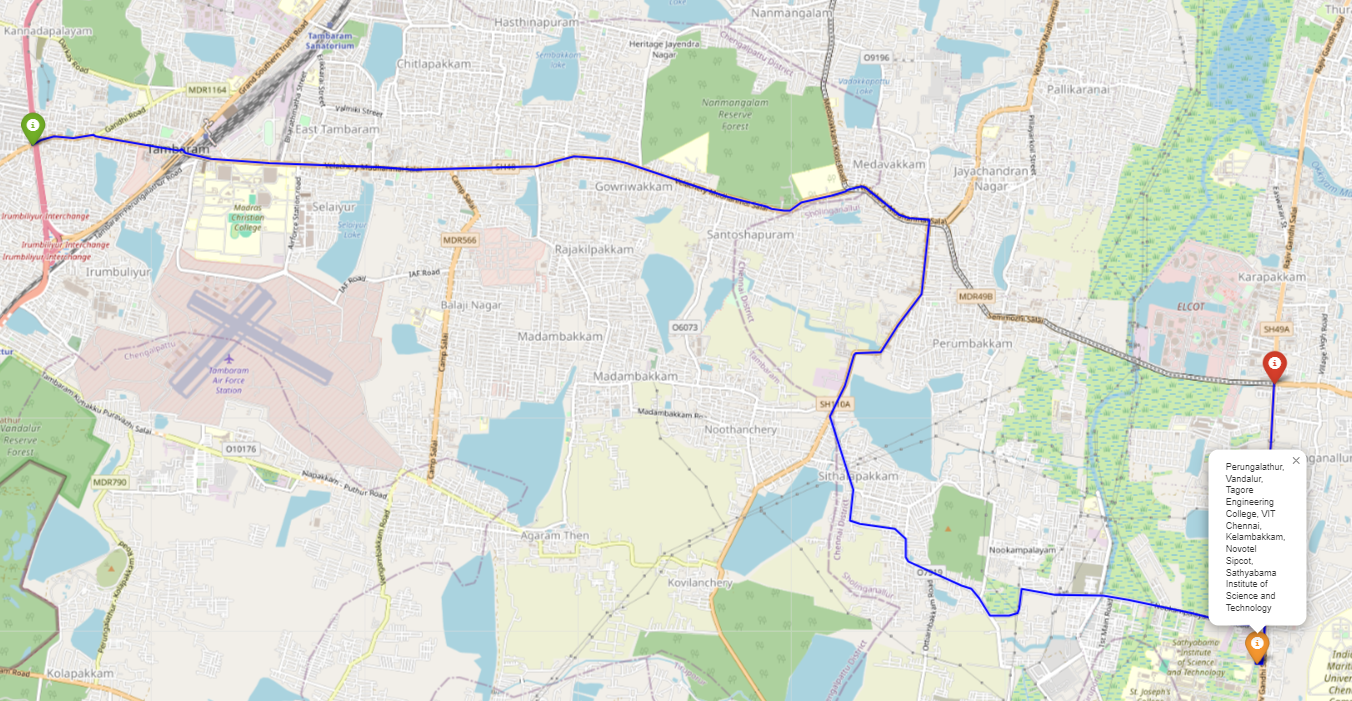}}
\captionsetup{justification=centering}
\caption{Final route between Tambaram and Sholingnallur with Intermediate stops}
\end{figure}

\section{Conclusion}
The model for route planning of metro network within a city is developed using modified ACO algorithm and by accessing \& processing GIS and other data. The existing rail route in the city  is determined \& compared to show the variations shown in the model's result with the original network. The intermediate stops along the new route are determined by considering the land elevation, population, important places \& nearby to important (point-of-interest) places, and the distance radius in which these conditions are to be considered. The implementation is completed and the results are visualized. The benefits of proposing this model in real-time is compared to that of factors involved during traditional method and the proposed model as shown a vast difference. There is potential for additional enhancement of the model in the future, in following ways:
\begin{enumerate}
    \item The model can be developed for route planning of other public mode of transportation like highways, long-distance bridges, waterways.
    \item It can be further applied on societies in different areas to provide a bigger impact on Global Sustainable Development Goals.
\end{enumerate}

\begin{IEEEbiographynophoto}{Hariram Sampath Kumar}
was born in Salem, Tamil Nadu, India in 2001. He received the B.E. degree in mechatronics engineering from Sona College of Technology (affiliated to Anna University, Chennai), Salem, Tamil Nadu in 2022. He is currently  pursuing the MTech degree in artificial intelligence at Amity University Noida, Uttar Pradesh, India.

His research interest includes development of field robots \& aerial robots, swarm intelligence, deep learning, and artificial intelligence for sustainable development.

Mr. Sampath Kumar has received Tamil Nadu State Council for Science \& Technology Grant under Student Projects Scheme 2021 – 2022.

\end{IEEEbiographynophoto}
\begin{IEEEbiographynophoto}{Archana Singh}
received the B.Sc degree in computer science from University of Delhi in 1997, Masters' degree in computer science and management from University of Pune, in 1999, MTech degree in computer engineering from IETE, New Delhi in 2008 and Ph.D. degree in data science focusing on data mining and machine learning from Amity University Noida in 2015.

She has 20 years of experience in academia and currently Professor \& Head - Department of Artificial Intelligence at Amity University Noida. Her research interests include artificial intelligence, machine learning, and image processing.
\end{IEEEbiographynophoto}
\begin{IEEEbiographynophoto}{Manish Kumar Ojha}
a faculty member at Amity University Noida, has extensive academic and professional experience in technology-driven engineering \& management. He received the BTech in Mechanical Engineering and MTech in Industrial Engineering and Management from IIT Kharagpur, India. He is currently pursuing the Ph.D. degree in Operations and Supply Chain Management from IIM Jammu, India. 

His research interests include Operations Research, Quantitative Methods, Supply Chain Management, and Optimization Techniques. He has also contributed significantly to academia, including publications in international journals and presentations at prestigious conferences. 

Mr. Ojha has also achieved notable achievements, including a DAAD Scholarship for a master's project at Technical University Dresden, Germany, and qualifying for CAT and GATE examinations.
\end{IEEEbiographynophoto}


\begin{thebibliography}{35}
\bibliographystyle{IEEEtran}

\bibitem{b5} J. L. Brown, “SDMtoolbox: a python-based GIS toolkit for landscape genetic, biogeographic and species distribution model analyses,” \textit{Methods in Ecology and Evolution}, vol. 5, no. 7, pp. 694–700, May 2014, doi: https://doi.org/10.1111/2041-210x.12200.
\bibitem{b6} J. Yu et al., “Particle swarm optimization based spatial location allocation of urban parks — A case study in Baoshan District, Shanghai, China,” Aug. 2014, doi: https://doi.org/10.1109/agro-geoinformatics.2014.6910575.
\bibitem{b7} L. H. Son, “Optimizing Municipal Solid Waste collection using Chaotic Particle Swarm Optimization in GIS based environments: A case study at Danang city, Vietnam,” \textit{Expert Systems with Applications}, vol. 41, no. 18, pp. 8062–8074, Dec. 2014, doi: https://doi.org/10.1016/j.eswa.2014.07.020.
\bibitem{b8} W. Zhang, Y. Zhou, W. Tian, and B. Hu, “GIS-based indoor pedestrian evacuation simulation combining the particle swarm optimization and the floor field model,” Jun. 2015, doi: https://doi.org/10.1109/geoinformatics.2015.7378551.

\bibitem{b10} T. S. T. Yusof, S. F. Toha, and H. Md. Yusof, “Path Planning for Visually Impaired People in an Unfamiliar Environment Using Particle Swarm Optimization,” \textit{Procedia Computer Science}, vol. 76, pp. 80–86, 2015, doi: https://doi.org/10.1016/j.procs.2015.12.281.
\bibitem{b11} H. Faris, I. Aljarah, S. Mirjalili, P. A. Castillo, and J. J. Merelo, “EvoloPy: An Open-source Nature-inspired Optimization Framework in Python,” \textit{Proceedings of the 8th International Joint Conference on Computational Intelligence}, 2016, doi: https://doi.org/10.5220/0006048201710177.

\bibitem{b15} Lucas Rodrigues Frank, Yan Mendes Ferreira, Eduardo Pagani Julio, F. Henrique, Bruno José Dembogurski, and Edelberto Franco Silva, “Multilayer Perceptron and Particle Swarm Optimization Applied to Traffic Flow Prediction on Smart Cities,” \textit{Lecture Notes in Computer Science}, pp. 35–47, Jan. 2019, doi: https://doi.org/10.1007/978-3-030-24305-0\_4.

\bibitem{b22} B. P. Kulkarni, S. Sai Krishna, K. Meenakshi, P. Kora and K. Swaraja, "Performance Analysis of Optimization Algorithms GA, PSO, and ABC based on DWT-SVD watermarking in OpenCV Python Environment," \textit{2020 International Conference for Emerging Technology (INCET)}, Belgaum, India, 2020, pp. 1-5, doi: 10.1109/INCET49848.2020.9154134.

\bibitem{b24} S. Jovanov and A. Naumoski, "A GIS-based Mapping of Mountain Peaks, Waterfalls and Mountain Lodges in North Macedonia," \textit{2020 4th International Symposium on Multidisciplinary Studies and Innovative Technologies (ISMSIT)}, Istanbul, Turkey, 2020, pp. 1-5, doi: 10.1109/ISMSIT50672.2020.9254627.
\bibitem{b25} N. Shatnawi, A. A. Al-Omari, and H. Al-Qudah, “Optimization of Bus Stops Locations Using GIS Techniques and Artificial Intelligence,” \textit{Procedia Manufacturing}, vol. 44, pp. 52–59, 2020, doi: https://doi.org/10.1016/j.promfg.2020.02.204.

\bibitem{b27} F. MiarNaeimi, G. Azizyan, and M. Rashki, “Horse herd optimization algorithm: A nature-inspired algorithm for high-dimensional optimization problems,” \textit{Knowledge-Based Systems}, vol. 213, p. 106711, Feb. 2021, doi: https://doi.org/10.1016/j.knosys.2020.106711.
\bibitem{b28} B. Abdollahzadeh, F. S. Gharehchopogh, and S. Mirjalili, “African vultures optimization algorithm: A new nature-inspired metaheuristic algorithm for global optimization problems,” \textit{Computers \& Industrial Engineering}, vol. 158, p. 107408, Aug. 2021, doi: https://doi.org/10.1016/j.cie.2021.107408.

\bibitem{b30} N. N. Samany, M. Sheybani, and S. Zlatanova, “Detection of safe areas in flood as emergency evacuation stations using modified particle swarm optimization with local search,” \textit{Applied Soft Computing}, vol. 111, p. 107681, Nov. 2021, doi: https://doi.org/10.1016/j.asoc.2021.107681.
\bibitem{b31} L. Abualigah, M. A. Elaziz, P. Sumari, Z. W. Geem, and A. H. Gandomi, “Reptile Search Algorithm (RSA): A nature-inspired meta-heuristic optimizer,” \textit{Expert Systems with Applications}, p. 116158, Nov. 2021, doi: https://doi.org/10.1016/j.eswa.2021.116158.
\bibitem{b32} S. N. Shorabeh, N. N. Samany, F. Minaei, H. K. Firozjaei, M. Homaee, and A. D. Boloorani, “A decision model based on decision tree and particle swarm optimization algorithms to identify optimal locations for solar power plants construction in Iran,” \textit{Renewable Energy}, vol. 187, pp. 56–67, Mar. 2022, doi: https://doi.org/10.1016/j.renene.2022.01.011.
\bibitem{b33} P. Shukla, S. Thakur, S. Arora and A. Wadhwa, "Nature-Inspired Algorithms Analysis on various Benchmark Functions using Python and Golang," \textit{2022 1st International Conference on Informatics (ICI)}, Noida, India, 2022, pp. 226-228, doi: 10.1109/ICI53355.2022.9786911.

\bibitem{b36} C. Zhong, G. Li, and Z. Meng, “Beluga whale optimization: A novel nature-inspired metaheuristic algorithm,” \textit{Knowledge-Based Systems}, p. 109215, Jun. 2022, doi: https://doi.org/10.1016/j.knosys.2022.109215.

\bibitem{b38} L. Belaiche, L. Kahloul, M. Houimli, S. Bousnane and S. Benharzallah, "Multi-Swarm-based Parallel Spider Monkey Optimization Algorithm," \textit{2022 International Conference on Advanced Aspects of Software Engineering (ICAASE)}, Constantine, Algeria, 2022, pp. 1-6, doi: 10.1109/ICAASE56196.2022.9931573.

\bibitem{b40} N. Eslami, S. Yazdani, M. Mirzaei, and E. Hadavandi, “Aphid–Ant Mutualism: A novel nature-inspired metaheuristic algorithm for solving optimization problems,” \textit{Mathematics and Computers in Simulation}, vol. 201, pp. 362–395, Nov. 2022, doi: https://doi.org/10.1016/j.matcom.2022.05.015.
\bibitem{b41} B. Abdollahzadeh, F. S. Gharehchopogh, N. Khodadadi, and S. Mirjalili, “Mountain Gazelle Optimizer: A new Nature-inspired Metaheuristic Algorithm for Global Optimization Problems,” \textit{Advances in Engineering Software}, vol. 174, p. 103282, Dec. 2022, doi: https://doi.org/10.1016/j.advengsoft.2022.103282.
\bibitem{b42} M. Abdel-Basset, R. Mohamed, M. Jameel, and M. Abouhawwash, “Nutcracker optimizer: A novel nature-inspired metaheuristic algorithm for global optimization and engineering design problems,” \textit{Knowledge-Based Systems}, p. 110248, Jan. 2023, doi: https://doi.org/10.1016/j.knosys.2022.110248.

\bibitem{b45} M. Han et al., “BasinMaker 3.0: A GIS toolbox for distributed watershed delineation of complex lake-river routing networks,” \textit{Environmental Modelling and Software}, vol. 164, pp. 105688–105688, Jun. 2023, doi: https://doi.org/10.1016/j.envsoft.2023.105688.

\bibitem{b47} Tarek Al Shammas, P. Gullón, O. Klein, and F. Escobar, “Development of a GIS-based walking route planner with integrated comfort walkability parameters,” \textit{Computers, Environment and Urban Systems}, vol. 103, pp. 101981–101981, Jul. 2023, doi: https://doi.org/10.1016/j.compenvurbsys.2023.101981.

\bibitem{b49} S. Vafadar, M. Rahimzadegan, and R. Asadi, “Evaluating the performance of machine learning methods and Geographic Information System (GIS) in identifying groundwater potential zones in Tehran-Karaj plain, Iran,” \textit{Journal of Hydrology}, vol. 624, p. 129952, Sep. 2023, doi: https://doi.org/10.1016/j.jhydrol.2023.129952.
\bibitem{b50} Adarsh Kumar Arya, Rajesh Katiyar, P. Senthil Kumar, A. Kapoor, Dan Bahadur Pal, and Gayathri Rangasamy, “A multi-objective model for optimizing hydrogen injected-high pressure natural gas pipeline networks,” \textit{International Journal of Hydrogen Energy}, vol. 48, no. 76, pp. 29699–29723, Sep. 2023, doi: https://doi.org/10.1016/j.ijhydene.2023.04.133.
\bibitem{b51} S. Durga, Padarbinda Samal, and Chinmoy Kumar Panigrahi, “Tyrannosaurus optimization algorithm: A new nature-inspired meta-heuristic algorithm for solving optimal control problems,” \textit{e-Prime}, vol. 5, pp. 100243–100243, Sep. 2023, doi: https://doi.org/10.1016/j.prime.2023.100243.
\bibitem{b52} G. Hu, Y. Guo, G. Wei, and Laith Abualigah, “Genghis Khan shark optimizer: A novel nature-inspired algorithm for engineering optimization,” \textit{Advanced Engineering Informatics}, vol. 58, pp. 102210–102210, Oct. 2023, doi: https://doi.org/10.1016/j.aei.2023.102210.

\bibitem{b54} O. Linsel and V. Bertsch, “A flexible approach to GIS based modelling of a global hydrogen transport system,” \textit{International Journal of Hydrogen Energy}, vol. 52, pp. 334–349, Jan. 2024, doi: https://doi.org/10.1016/j.ijhydene.2023.08.199.

\bibitem{b57} E.-S. M. El-kenawy, Nima Khodadadi, Seyedali Mirjalili, A. A. Abdelhamid, M. M. Eid, and A. Ibrahim, “Greylag Goose Optimization: Nature-inspired optimization algorithm,” \textit{Expert Systems with Applications}, vol. 238, pp. 122147–122147, Mar. 2024, doi: https://doi.org/10.1016/j.eswa.2023.122147.
\bibitem{b58} M. Ahmed, Mohd Herwan Sulaiman, Ahmad Johari Mohamad, and M. Rahman, “Gooseneck barnacle optimization algorithm: A novel nature inspired optimization theory and application,” \textit{Mathematics and Computers in Simulation}, vol. 218, pp. 248–265, Apr. 2024, doi: https://doi.org/10.1016/j.matcom.2023.10.006.
\bibitem{b59} M. Han, Z. Du, K. F. Yuen, H. Zhu, Y. Li, and Q. Yuan, “Walrus optimizer: A novel nature-inspired metaheuristic algorithm,” \textit{Expert Systems with Applications}, vol. 239, p. 122413, Apr. 2024, doi: https://doi.org/10.1016/j.eswa.2023.122413.
\bibitem{b60} H. Sampath Kumar, A. Singh, M. K. Ojha, "Artificial Intelligence Based Navigation in Quasi Structured Environment", unpublished.
\bibitem{b61} N. R. Haddaway, M. J. Page, C. C. Pritchard, and L. A. McGuinness, “PRISMA2020: An R package and Shiny app for producing PRISMA 2020‐compliant flow diagrams, with interactivity for optimised digital transparency and Open Synthesis,” \textit{Campbell Systematic Reviews}, vol. 18, no. 2, Mar. 2022, Available: https://onlinelibrary.wiley.com/doi/full/10.1002/cl2.1230

\end{thebibliography}
\end{document}